\documentclass[10pt,twocolumn,letterpaper]{article}

\usepackage[pagenumbers]{iccv} %

\newcommand{\modelname}{\mbox{{St4RTrack}}\xspace}

\newcommand{\dus}[1]{\mbox{DUSt3R}}
\newcommand{\mon}[1]{MonST3R}
\newcommand{\mas}[1]{MASt3R}

\newcommand{\qheading}[1]{\noindent\mbox{\textbf{#1}}}

\newif\ifdrafting
\draftingtrue %
\ifdrafting
    \newcommand{\todo}[1]{{\leavevmode\color[rgb]{1,0,0}[TODO: #1]}}
    \newcommand{\jz}[1]{{\leavevmode\color[rgb]{0,0.8,0.2}[Junyi: #1]}}
    \newcommand{\hf}[1]{{\leavevmode\color[rgb]{0.6,0.,0.8}[Haven: #1]}}
    \newcommand{\ak}[1]{{\leavevmode\color[rgb]{0,0.6,0.8}[AK: #1]}}
    \newcommand{\TD}[1]{{\leavevmode\color[rgb]{0,0.8,0.9}[TD: #1]}}
    \newcommand{\yy}[1]{{\leavevmode\color[rgb]{0.2,0.4,0.8}[Yufei: #1]}}
    \newcommand{\mb}[1]{{\leavevmode\color[rgb]{0.8,0.2,0.1}[MJB: #1]}}

\else
    \newcommand{\todo}[1]{}
    \newcommand{\jz}[1]{}
    \newcommand{\hf}[1]{}
    \newcommand{\yy}[1]{}
    \newcommand{\ak}[1]{}
    \newcommand{\mb}[1]{}
    \newcommand{\TD}[1]{}
\fi

\definecolor{iccvblue}{rgb}{0.21,0.49,0.74}
\usepackage[pagebackref,breaklinks,colorlinks,allcolors=iccvblue]{hyperref}
\usepackage{amsmath}
\usepackage{mathtools}
\usepackage{dsfont}

\usepackage{multirow}

\DeclareMathOperator*{\argmin}{argmin}
\def\paperID{4606} %
\def\confName{ICCV}
\def\confYear{2025}

\title{St4RTrack: Simultaneous 4D Reconstruction and Tracking in the World}

\author{
Haiwen Feng$^{1,2\ *}$ \quad
Junyi Zhang$^{1\ *}$ \quad
Qianqian Wang$^{1}$ \quad
Yufei Ye$^{3}$ \quad
Pengcheng Yu$^{2}$ \\
Michael J. Black$^{2}$ \quad
Trevor Darrell$^{1}$ \quad
Angjoo Kanazawa$^{1}$ \\
\vspace{-0.5em}
\\
$^1$UC Berkeley \quad
$^2$Max Planck Institute for Intelligent Systems \quad
$^3$Stanford University \\
}

\definecolor{deepred}{RGB}{189,0,0}   %
\definecolor{deepblue}{RGB}{0,0,189}    %
\definecolor{deeppurple}{RGB}{180,80,180} %
\definecolor{deepgreen}{RGB}{80,200,80}   %
\definecolor{deepyellow}{RGB}{204,153,0}   %

\begin{document}
\twocolumn[{
\maketitle

\vspace{-2em}
\begin{center}
    \begin{minipage}{\linewidth}
        \centering
        \includegraphics[width=\textwidth]{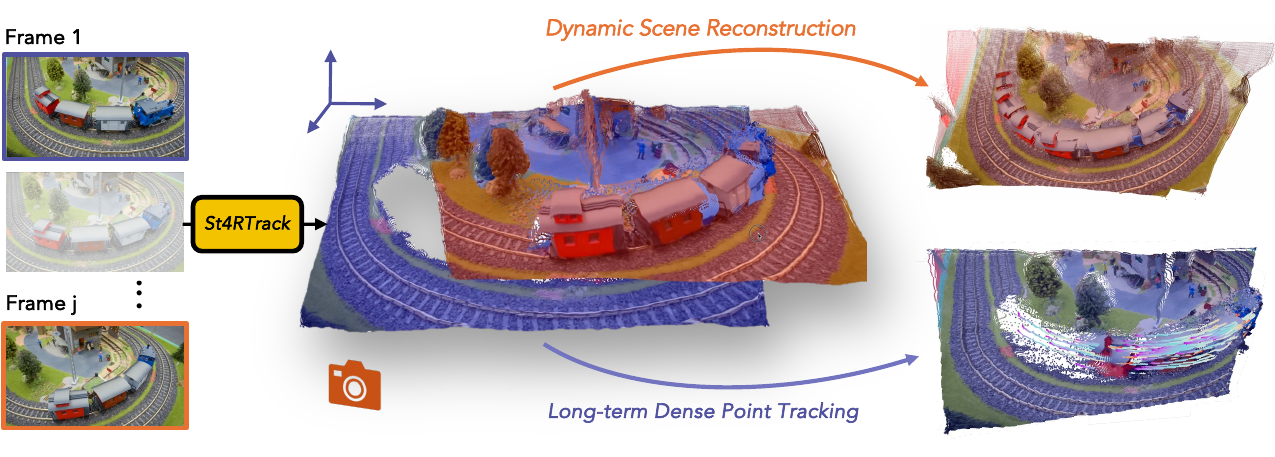}
        \vspace{-1.5em}
        \captionof{figure}{\textbf{\modelname:} Given an RGB video capturing dynamic scenes, \modelname simultaneously tracks the points from the initial frame (\textcolor{deepblue}{visualized in purple}) and reconstructs the geometry of the subsequent frames (\textcolor{deepred}{in orange}) in a consistent world coordinate frame. 
        \modelname is a \textit{feed-forward} framework that takes a pair of images as input and outputs two pointmaps in the world frame, as the visualization shown in the middle. 
        By iteratively processing the first frame paired with each subsequent frame, \modelname achieves simultaneous tracking (right bottom) and reconstruction (right top) for the entire video.
        Interactive results on our webpage: \url{https://St4RTrack.github.io/}.
        }
        \vspace{1.0em}
        \label{fig:teaser}
        
    \end{minipage}
\end{center}
}]
\renewcommand{\thefootnote}{}
\footnotetext{$^*$Equal contribution, listed alphabetically.}

\begin{abstract}
Dynamic 3D reconstruction and point tracking in videos are typically treated as separate tasks, despite their deep connection. We propose \modelname, a feed-forward framework that simultaneously reconstructs and tracks dynamic video content in a world coordinate frame from RGB inputs. This is achieved by predicting two appropriately defined pointmaps for a pair of frames captured at different moments. Specifically, we predict both pointmaps \emph{at the same moment, in the same world}, capturing both static and dynamic scene geometry while maintaining 3D correspondences. Chaining these predictions through the video sequence with respect to a reference frame naturally computes long-range correspondences, effectively combining 3D reconstruction with 3D tracking. Unlike prior methods that rely heavily on 4D ground truth supervision, we employ a novel 
 adaptation scheme based on a reprojection loss. We establish a new extensive benchmark for world-frame reconstruction and tracking, demonstrating the effectiveness and efficiency of our unified, data-driven framework. Our code, model, and benchmark will be released.

\end{abstract}
    
\section{Introduction}
\label{sec:intro}
When asked about the three most important problems in computer vision, Takeo Kanade replied, “Correspondence, Correspondence, Correspondence!” This remark is especially pertinent to multi-view 3D reconstruction, where 3D geometry and 2D correspondence are two sides of the same coin; that is, a 3D point in the physical world naturally brings 2D correspondence across its projections in different views, and conversely, corresponded 2D points across views reconstruct the same 3D point after triangulation. This synergy between 3D geometry and 2D correspondence has long formed the foundation of multi-view geometry~\cite{hartley2003multiple}. However, when the scene becomes dynamic, this synergy appears to vanish, as existing methods—particularly the recent data-driven ones—tend to treat dynamic reconstruction~\cite{zhang2025monst3r,wang2025cut3r,karaev2023dynamicstereo} and correspondence~\cite{ngo2025delta,karaev23cotracker,teed2020raftrecurrentallpairsfield} as separate, disconnected tasks. We argue that this is a missed opportunity; the synergy between 3D reconstruction and 2D correspondence is not lost in dynamic scenes—it simply requires an additional element: understanding how the scene content evolves over time. This evolution is captured by 3D motion estimation (\ie, dense 3D point tracking), which, when computed across a sequence, effectively explains scene motion. Once tracking accounts for scene dynamics, the problem effectively reduces to the rigid case, where the natural interplay between reconstruction and correspondence can once again be leveraged.

We propose \modelname, a learning framework that unifies reconstruction and tracking directly from RGB video frames. \modelname simultaneously reconstructs and tracks dynamic video content in a single consistent world, achieving world-frame 3D tracking, as demonstrated in Fig.~\ref{fig:teaser}. Tracking in the world frame decouples the scene  and the camera motion, essential for domains where both the camera and content are in motion. Our approach also reconstructs the 3D geometry of observed image content for both static and dynamic portions of the scene. \modelname directly predicts their reconstruction and tracking in the world without requiring an additional alignment stage.

\begin{figure}   
\includegraphics[width=\linewidth]{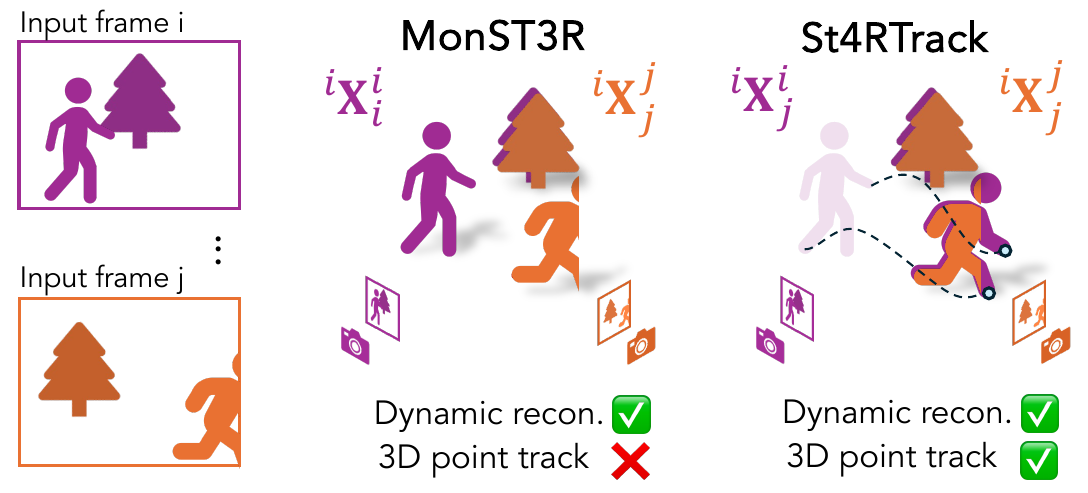}
    \captionof{figure}{\textbf{Pointmap Comparison of MonST3R and St4RTrack.} 
    Given two input frames, MonST3R handles \textit{dynamic} scenes by reconstructing both pointmaps in their own timestamp. St4RTrack  predicts where the points in the first frame move in the second frame, and reconstructs the geometry of the second frame. More details of the representation definition are introduced in~\cref{sec:representation}.
}
    \label{fig: concept}
\end{figure}

Our key insight stems from the observation that a feed-forward static 3D reconstruction method, \dus{}~\cite{Wang2023DUSt3RG3}, can be adapted to dynamic scenes simply by changing the pointmaps’ annotation~\cite{zhang2025monst3r}. Building on this, we reexamine the pointmaps definition in the 4D scenario and opt to simply redefine its geometric interpretation for both reconstruction and tracking, as illustrated in Fig.~\ref{fig: concept}. 
Concretely, we achieve it by predicting two pointmaps \emph{at the same timestamp} and \emph{in the same world} from a pair of image frames depicting two different timestamps. More specific, given images $(\mathbf{I}_i, \mathbf{I}_j)$, both pointmaps are predicted in the coordinate frame of $\mathbf{I}_i$, but at the time specified by $\mathbf{I}_j$. Our method is realized through a feed-forward network comprising of two branches: the \emph{reconstruction branch} reconstructs the content of $\mathbf{I}_j$ in the $\mathbf{I}_i$ coordinate frame; and the \emph{tracking branch}, which reconstructs the content of $\mathbf{I}_i$ in the $\mathbf{I}_i$ (its own) coordinate frame, \textit{but} at the time indicated by $\mathbf{I}_j$. Essentially, the tracking branch predicts how the scene content in $\mathbf{I}_i$ evolves to match the moment captured in $\mathbf{I}_j$. This is enabled through a DUSt3R-like dual cross-attention mechanism, where the tracking branch relies on the reconstruction branch to decide how to move points.  
This minimal change proves sufficient for unifying both dynamic reconstruction and 3D point tracking in the world coordinate system.

Furthermore, unlike existing methods~\cite{Wang2023DUSt3RG3,zhang2025monst3r,wang2025cut3r} that rely solely on 4D supervision, our approach unlocks 4D reconstruction training on in-the-wild videos via reprojection loss \emph{without} 4D supervision. This is possible because \modelname simultaneously establishes camera parameters, 3D geometry and motion. 
Specifically, based on the outputs of the reconstruction branch, the camera parameters for \(\mathbf{I}_j\) can be differentiably computed via PnP. Using these cameras, the pointmap of \(\mathbf{I}_i\) is projected into the \(j\)-th frame, enabling training with a reprojection loss that leverages 2D correspondences and monocular depth predictions from off-the-shelf approaches~\cite{wang2024moge,karaev2024cotracker3}. 
Consequently, the monocular supervisions facilitate effective \emph{test-time adaptation} of \modelname to in-the-wild videos, which can differ substantially from the synthetic data used during pretraining.

While prior 3D point tracking benchmarks focus on camera coordinate frames~\cite{xiao2024spatialtracker}, our approach enables world-frame 3D tracking. To evaluate this capability, we establish a novel benchmark, WorldTrack, for both tracking and reconstruction in the world coordinate system. 
We find that our unified method outperforms the strong baselines that combine several pieces on each individual task. Furthermore, we show that our feedforward results can be improved via test-time adaptation. 
We believe this is a step towards a unified task-agnostic 4D perception system that can be trained on a large-scale video. Our code, model, and benchmark will be released.

\section{Related Works}
\label{sec:related}

\paragraph{Camera Estimation and Scene Reconstruction.}
Jointly estimating camera motion and scene geometry has been studied for decades, often in the context of Structure from Motion (SfM)~\cite{schonberger2016structure,agarwal2011building,seitz2006comparison,wang2024vggsfm} or Simultaneous Localization and Mapping (SLAM)~\cite{durrant2006simultaneous,davison2007monoslam,mur2015orb,wang2017deepvo,teed2021droid}. However, these methods are primarily designed for static scenes and typically do not model dynamic scene content. Recent advances in learning-based monocular and video depth estimation methods~\cite{ranftl2021vision,yang2024depth,bhat2023zoedepth,piccinelli2024unidepth} have opened new opportunities to reconstruct dynamic scenes. Notably, R-CVD~\cite{kopf2021robust}, 
CasualSAM~\cite{zhang2022casualsam} and MegaSAM~\cite{li2024megasam} jointly optimize camera parameters and per-frame dense depth maps leveraging monocular depth priors, producing consistent depth estimates for dynamic objects and accurate camera parameters even in challenging cases with minimal camera parallax. 
Another notable recent method, DUSt3R~\cite{Wang2023DUSt3RG3}, introduces a two-pointmap representation that enables joint estimation of camera motion and scene geometry of a pair of images. While DUSt3R itself primarily focuses on reconstructing static scenes, 
follow-up effort such as MonST3R~\cite{zhang2025monst3r}, demonstrate that this formulation can also effectively handle dynamic scenes with minimal modification on supervisions. Despite these advances, none of the aforementioned methods explicitly estimate 3D scene motion, meaning they do not track the movement of individual 3D points over time. In contrast, our method simultaneously performs joint reconstruction and tracking for dynamic scenes.

\vspace{0.5em}
\qheading{2D/3D Point Tracking.}
Tracking pixel motion over time is a fundamental problem in computer vision. Optical/Scene flow methods~\cite{horn1981determining,lucas1981iterative,black1993framework,sun2018pwc, teed2021raft,teed2020raftrecurrentallpairsfield,hur2020self, vedula1999three} produce dense 2D/3D motion vectors but are inherently short-ranged, struggling with large displacements and occlusions. While long-range point tracking~\cite{1641022,rubinstein2012towards} has been studied for decades, it has recently been revitalized via supervised learning~\cite{harley2022particle,karaev23cotracker,karaev2024cotracker3,doersch2022tap,doersch2023tapir}, enabling more robust tracking over extended time periods and overcoming these limitations.
However, these methods still produce only 2D pixel trajectories.
More recently, several works~\cite{xiao2024spatialtracker,ngo2025delta} achieve 3D tracking by lifting points into 3D space using monocular depth priors and performing tracking in 3D. While closely related to our approach, these methods still operate in the camera frame space, meaning they lack camera motion estimation and do not explicitly separate scene motion from camera motion. In contrast, our method jointly estimates disentangled camera and scene motion, enabling world-space tracking for a more complete understanding of 3D scene dynamics.

\vspace{0.5em}
\qheading{Joint Dynamic Reconstruction and Tracking.} 
While traditional non-rigid SfMs have studied joint tracking and reconstruction~\cite{novotny2019c3dpo,bregler2000recovering,dai2014simple,fragkiadaki2014grouping,zhou2016sparseness} from 2D correspondences, jointly optimizing them from raw videos is highly challenging and typically requires multi-view synchronized videos as input~\cite{luiten2023d,li2022neural,wu20244d,fridovich2023k,park2021hypernerf}. 
With recent advances in neural rendering and data-driven geometric priors, reconstructing and rendering dynamic scenes from monocular videos became possible. However, as Gao et al.~\cite{gao2022monocular} point out, many methods~\cite{pumarola2021d,yang2024deformable} focus on ``teleporting'' input data, which are effectively multi-view and not representative of real-world videos. In addition, since the main focus is view synthesis,  motion estimation serves a secondary role in facilitating information fusion between neighboring frames~\cite{li2021neural,li2023dynibar}.  More recently, several works~\cite{liu2024modgsdynamicgaussiansplatting, lei2024mosca, wang2024shapeofmotion} focus on jointly recovering camera parameters, persistent scene geometry, and long-range 3D tracks from single, causally captured videos.  However, these methods take off-the-shelf priors as given and design per-video optimization techniques that optimize a representation from scratch. 
Most recently, Stereo4D~\cite{jin2024stereo4d}—a concurrent effort to our work—proposed a pipeline for crafting a real-world 4D tracking dataset using internet stereo videos, enabling the supervised regression of 3D trajectories and geometries between frames. 
In contrast, we propose a feed-forward method that simultaneously performs reconstruction and tracking, while the same architecture also supports test-time adaptation on unlabeled videos to approach the high quality of optimization-based methods.

\begin{figure*}[htbp]
  \centering
  \includegraphics[width=1\textwidth]{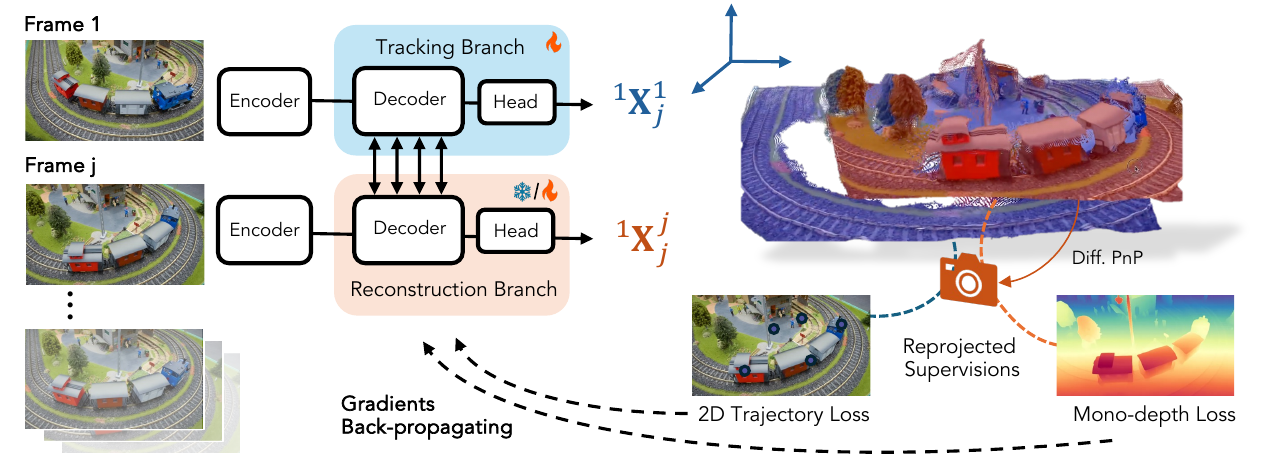} %
  \caption{\textbf{Overview of \modelname.} Given frame $1$ and frame $j$ as input, the tracking branch outputs ${}^{\textcolor{deepyellow}{1}}\mathbf{X}^{\textcolor{deeppurple}{1}}_{\textcolor{deepgreen}{j}}$, 
  the pointmap that corresponds to \textcolor{deeppurple}{observed content} of the first frame at \textcolor{deepgreen}{timestep} $j$ in its own \textcolor{deepyellow}{camera coordinate} (\ie world coordinate); 
  the reconstruction branch outputs ${}^{\textcolor{deepyellow}{1}}\mathbf{X}^{\textcolor{deeppurple}{j}}_{\textcolor{deepgreen}{j}}$, the pointmap of the \textcolor{deeppurple}{content} in frame $j$ at its own \textcolor{deepgreen}{timestamp} in the world \textcolor{deepyellow}{coordinate}.
  To adapt to new videos without any 4D labels, the camera is computed via differentiable PnP from the pointmap, enabling reprojected supervision signals (e.g., 2D trajectories and monocular depth). We finetune both branches during training (Sec.~\ref{sec:method}) with synthetic data, and when adapting to a new video (Sec.~\ref{sec:adapt}), only the tracking branch is fine-tuned using these reprojected supervision signals.
  }
  \label{fig:pipeline}
\end{figure*}

\section{Simultaneous Reconstruction and Tracking}

We present a framework that simultaneously reconstructs and tracks dynamic video content in 3D within a single consistent world coordinate frame. 
The core idea is simple yet powerful: reconstructing and tracking can both be achieved by predicting two appropriately defined pointmaps, where both pointmaps reconstruct the scene content observed in each image \textit{at the same timestamp} in a \textit{consistent coordinate system}. 
This enables simultaneous reconstruction of both dynamic and static contents, while tracking across a sequence of image frames in a video. 
Since all geometry, camera, and motion (\ie 3D correspondence over time) can be derived from the representation, it can be adapted to videos without any explicit 4D supervision. 
Below, we discuss the main insight, how \modelname compares to prior works. Then, we discuss the details of the model and how it can be trained and adapted to videos.

\subsection{Unified 4D Representation of \modelname} 
\label{sec:representation}
Given two images \(\mathbf{I}_i, \mathbf{I}_j\) with dynamic content (see Figure~\ref{fig: concept}), how can one devise a single feedforward approach that simultaneously performs reconstruction and tracking? We argue that the underlying representation must (1) capture camera motion to establish a world coordinate frame, (2) reconstruct the 3D geometry of all observed points, and (3) estimate 3D motion that maintain explicit correspondence over time. In this work, we show that just two properly defined pointmaps suffice to fulfill these requirements.

\vspace{0.5em}
\qheading{Time-Dependent Pointmap.}
A pointmap representation assumes that each pixel in an image \(\mathbf{I}\) of shape \({H \times W}\) is associated with a corresponding 3D point, forming a pointmap \(\mathbf{X} \in \mathbb{R}^{H \times W \times 3}\). For the case of static scenes, \dus{}~\cite{Wang2023DUSt3RG3} only considers two factors of pointmaps—(1) the source frame of the content and (2) the camera coordinate system in which the points are expressed. However, this definition is insufficient for modeling the dynamic scenario of monocular video. 
To address this, we introduce a previously overlooked yet decisive factor: \textit{time}.

Specifically, we define a time-dependent pointmap that encodes the 3D positions of the scene points \emph{in a chosen (world) coordinate system} \emph{at a specific timestamp}. For clarity, we denote this representation as \({}^{\textcolor{deepyellow}{a}}\mathbf{X}^{\textcolor{deeppurple}{b}}_{\textcolor{deepgreen}{t}}\), which denotes the 3D pointmap of \textcolor{deeppurple}{physical content} from frame \(b\), at \textcolor{deepgreen}{time} \(t\), expressed in the \textcolor{deepyellow}{coordinate system} established by frame \(a\).” 
For example, \({}^{i}\mathbf{X}^{i}_{j}\) represents the geometry originally seen in frame \(i\) in frame \(i\)’s own coordinate system, but described at timestamp \(j\).
The time-dependency is achieved by construction without explicit timestamp conditioning, as described in the next section.

\vspace{0.5em}
\qheading{Unified 4D Modeling.}
\modelname learns a function $f$ that maps two images $\mathbf{I}_i, \mathbf{I}_j$, captured at timestamp $i$ and $j$, into two pointmaps:
\begin{equation}
    f_\theta(\mathbf{I}_i, \mathbf{I}_j) = {}^{i}\mathbf{X}^{i}_{j},  {}^{i}\mathbf{X}^{j}_{j}.
    \label{eq1}
\end{equation} 
The second image \(\mathbf{I}_j\) is reconstructed as the pointmap \({}^{i}\mathbf{X}^{j}_{j}\) in the first image $\mathbf{I}_i$'s coordinate frame. Meanwhile, it predicts \({}^{i}\mathbf{X}^{i}_{j}\), representing the 3D motion of how the content from the first image $\mathbf{I}_i$ moves at timestamp \(j\). 
Thus, both geometry and motion (tracking) are estimated from this unified prediction.

To handle a full video consisting of $T$ frames, we perform tracking and reconstruction by always selecting the first frame as the anchor frame $\mathbf{I}_i$. Each subsequent frame $\mathbf{I}_j$ is then paired with this initial frame, ensuring that every new frame is consistently aligned to the coordinate system of the first frame. Specifically, \(\{ f(\mathbf{I}_1, \mathbf{I}_1), f(\mathbf{I}_1, \mathbf{I}_2), \ldots, f(\mathbf{I}_1, \mathbf{I}_T)\}\) are computed in the same reference, $\mathbf{I}_1$, which naturally serves as the world coordinate frame. Thus, world-frame 3D tracking is achieved by explicitly following how points observed in \(\mathbf{I}_1\) are placed throughout the sequence, \(\{{}^{1}\mathbf{X}^{1}_{1}, {}^{1}\mathbf{X}^{1}_{2}, \dots, {}^{1}\mathbf{X}^{1}_{T}\}\), while the world frame dynamic reconstruction is obtained by the paired geometry estimation per-frame, \(\{{}^{1}\mathbf{X}^{1}_{1}, {}^{1}\mathbf{X}^{2}_{2}, \dots, {}^{1}\mathbf{X}^{T}_{T}\}\).

\vspace{0.5em}
\qheading{Relation to prior works.}
Our formulation of the 4D modeling generalizes prior works in a unified framework. DUSt3R reconstructs and establishes correspondences but is limited to rigid scenes, as correspondence and reconstruction are dual tasks for static scenes.
With this perspective, one can see that if there is no dynamic component (i.e. frozen moment in time or rigid scenes), our formulation is equivalent to \dus{}, where both images share the same timestamp $t=i$:
\begin{equation}
    f_\theta(\mathbf{I}_i, \mathbf{I}_j) = {}^{i}\mathbf{X}^{i}_{i},  {}^{i}\mathbf{X}^{j}_{i}.
\end{equation}

In a static world, 3D reconstruction from two pointmaps inherently yields the correspondences between them, allowing synergy to arise naturally. However, when objects or the scene are in motion, the dynamic component appears differently in different frames, it becomes crucial to account for 3D scene motion to preserve this synergy. \modelname addresses this challenge by predicting the 3D content from the first image at future timestamps.

In the same framework, we see that \mon{}, the dynamic follow-up of \dus{} can be expressed as such: 
\begin{equation}
    f_\theta(\mathbf{I}_i, \mathbf{I}_j) = {}^{i}\mathbf{X}^{i}_{i},  {}^{i}\mathbf{X}^{j}_{j},
\end{equation}
where each image's 3D geometry is reconstructed in its timestamp, such that the dynamic contents separately align with their frame inputs. While it's sufficient for obtaining dynamic scene geometry, there is no temporal correspondence being established, as illustrated in~\cref{fig: concept}. Furthermore, both \dus{} and \mon{} compute the pairwise graphs and perform global alignment.

Since we always designate the first frame as the reference for tracking, the world coordinates are consistently established by the first frame. For simplicity, we omit the explicit notation of the world coordinate in subsequent equations and paragraphs, \ie, $\mathbf{X}^{i}_{j}:= {}^{i}\mathbf{X}^{i}_{j}$.

\subsection{Joint Learning of Tracking and Reconstruction}
\label{sec:method}
In this section, we describe how our framework implements equation~\ref{eq1} within a pair-wise framework as \dus{}. For each pair of frames, \(\mathbf{I}_1\) and \(\mathbf{I}_j\), we first encode them into token representations using a ViT encoder, then process these tokens through a siamese transformer decoder. The decoder sequentially applies self-attention (allowing tokens within each frame to interact), followed by cross-attention (enabling tokens from one frame to attend to tokens in the other), and finally passes the tokens through an MLP. This continuous information flow between the two branches is crucial for generating spatial-aligned 3D pointmaps in a shared coordinate system, as illustrated in Fig.~\ref{fig:pipeline}.

Our siamese architecture processes two input views concurrently and generates two 3D pointmaps that are expressed in a common reference frame established by the first view. Although the two branches share the same architectural structure, they serve distinct purposes:
\begin{itemize}
    \item \textbf{Tracking branch} predicts the pointmap \(\mathbf{X}^{1}_{j}\), which represents the geometry of the first frame at timestamp \(j\) in the first frame’s coordinates (\ie, the world coordinates).
    \item \textbf{Reconstruction branch} predicts the pointmap \(\mathbf{X}^{j}_{j}\), which represents the geometry of frame \(j\) at its own timestamp, also expressed in the first frame’s camera coordinates.
\end{itemize}
Since this architecture is exactly the same as proposed by \dus{} and subsequently adopted by \mon{}, with the only difference being the output paired pointmaps (Eq.1-3), our network can be initialized with pretrained 3D knowledge from either \dus{} or \mon{}. 

\vspace{0.5em}
\qheading{Pretraining with 4D Synthetic Data.}
Our proposed representation requires specialized supervision for the Tracking Branch—namely, ensuring that the pointmap from the first frame is correctly positioned in the world across all frames. Achieving this necessitates complete 4D information of the dynamic scene. Therefore, we leverage existing 4D synthetic datasets~\cite{zheng2023pointodyssey, karaev2023dynamicstereo} that provide both the 3D geometry and motion of the rendered content. Specifically, for each dataset, we use the scene mesh vertices (expressed in world coordinates) to provide sparse, masked supervision for the Tracking Branch pointmap representation, and employ per-frame depth maps and camera ground-truth to supervise the Reconstruction Branch. For this fully supervised training process, we use the objectives from \dus{}.
We initialize our dual-branch transformer with weights from MASt3R~\cite{leroy2024grounding}, a \dus{} variant that has been adapted for 2D correspondence learning.
Additional details regarding the 4D synthetic datasets are provided in~\cref{sec:exp_details}. 

\subsection{Adapt to Any Video without 4D Label}
\label{sec:adapt}
While the synthetic datasets are small-scale and unrealistic, they are sufficient for our network to learn the newly proposed representations. However, fully supervised training on these datasets presents two key limitations: 1) The 4D synthetic data is limited in scale and does not encompass the full range of motion and geometry present in real-world dynamic scenes; 2) Our proposed pointmap representation requires the capability to freely move the pointmap within the world coordinates—a departure from conventional pixel-aligned geometry predictions, making small-scale training insufficient for achieving fine-grained predictions.
These limitations motivate us to further leverage the 3D geometry and motion inherent in the \modelname framework to perform domain adaptation on any video without 4D labels. Specifically, we first show how we can derive camera parameters differentially, and with which we can design reprojected 2D trajectory loss and monocular depth loss to supervise the network. 

\vspace{0.5em}
\qheading{Solving Camera Parameters.}
The intrinsic matrix \(\mathbf{K}\) is first estimated from the tracking branch's first-frame pointmap prediction, following \dus{}~\cite{Wang2023DUSt3RG3}. In this process, the principal point is assumed to be centered, and pixels are treated as square. The focal length is assumed static across frames and estimated using a fast iterative solver based on the Weiszfeld algorithm~\cite{weiszfeld1937point}.
Next, the extrinsic parameters \(\mathbf{P}^j = [\mathbf{R}^j|\mathbf{T}^j]\) for each frame \(j\) are derived using the “reconstruction” pointmap \(\mathbf{X}^j_j\). Specifically, each pixel \(\mathbf{x}^{j,n}\) in frame \(j\) is associated with a 3D coordinate \(\mathbf{X}^{j,n}_{j}\) in the shared world coordinate system (established by the first camera), thus forming 2D-to-3D correspondences.
We could then solve for \(\mathbf{R}^j\) and \(\mathbf{T}^j\) via a Perspective-\(n\)-Points (PnP)~\cite{lepetit2009ep} solver with RANSAC~\cite{fischler1981random} for outlier rejection:
 \begin{equation}
 \mathbf{R}^j, \mathbf{T}^j = \argmin_{\mathbf{R}, \mathbf{T}} \sum_{n \in \mathcal{I}_j} 
 \Bigl\|\mathbf{x}^{j,n} - \pi\bigl(\mathbf{K}\,(\mathbf{R}\,\mathbf{X}^{j,n}_{j} + \mathbf{T})\bigr)\Bigr\|^2,
 \end{equation}
 where \(\pi(\cdot)\) is the projection \((x,y,z)\rightarrow(x/z,y/z)\).  
 
 For differentiability, we adopt a derivative-based Gauss-Newton solver following~\cite{chen2023epro}, ensuring that gradients from the reprojection loss can adjust both the camera pose and the 3D pointmaps. Further details are provided in~\cref{sec:diff_cam}.

\vspace{0.5em}
\qheading{Reprojection Loss.}
With the camera pose of frame \(j\) derived, the \emph{tracking} pointmap \(\mathbf{X}^{1}_{j}\) and \emph{reconstruction} pointmap \(\mathbf{X}^{j}_{j}\) can be transformed from the world coordinate system into the camera coordinate system of frame \(j\). This transformation enables self-supervised training by enforcing two types of consistency: (1) 2D correspondence consistency, which aligns the projected 2D tracks (from \(\mathbf{X}^{1}_{j}\)) with the pseudo-ground truth tracking from CoTracker~\cite{karaev23cotracker,karaev2024cotracker3}, and (2) geometric consistency, which aligns the scale-invariant depth (from \(\mathbf{X}^{j}_{j}\)) with the pseudo-ground truth monocular depth from MoGe~\cite{wang2024moge}.

More specifically, given the estimated camera pose \((\mathbf{R}^j, \mathbf{T}^j)\) from frame \(j\) and the tracking pointmap \(\mathbf{X}^{1}_{j}\), we reproject these 3D points into the image plane of frame \(j\):
\begin{equation}
\hat{\mathbf{x}}^{j,n} = \pi \bigl( \mathbf{K} \bigl(\mathbf{R}^j\,\mathbf{X}^{1,n}_{j} + \mathbf{T}^j\bigr) \bigr).
\end{equation}
These reprojected points serve as the predicted 2D tracks and are compared with the pseudo-ground truth tracking points \(\mathbf{x}_{\text{trk}}^{j,n}\) from CoTracker3~\cite{karaev2024cotracker3}.

To mitigate minor focal inaccuracies that may induce scaling shifts, the reprojection loss is computed in a scale-invariant manner. Let \(\mathbf{p}_n = \hat{\mathbf{x}}^{j,n}\) and \(\mathbf{g}_n = \mathbf{x}_{\text{trk}}^{j,n}\) for \(n = 1,\dots,N\), and denote the image center by \(c\). The scale factor and adjusted predictions are computed together as:
\begin{equation}
\hat{\mathbf{p}}_n = (\mathbf{p}_n - c)\, s + c, \quad s = \frac{1}{N}\sum_{n=1}^{N} \frac{\|\mathbf{g}_n - c\|_2}{\|\mathbf{p}_n - c\|_2}.
\end{equation}
Then, the scale-invariant 2D reprojection loss is defined as
\begin{equation}
\mathcal{L}_{\text{traj}} 
= \frac{1}{N} \sum_{n \in \mathcal{I}_j} \|\hat{\mathbf{p}}_n - \mathbf{g}_n\|^2.
\end{equation}

Similarly, to enforce geometric consistency with the mono-depth predictions from MoGe~\cite{wang2024moge}, we use the reconstruction pointmap \(\mathbf{X}^{j}_{j}\) that transformed to frame $j$'s camera coordinate. The depth of each transformed 3D point is
\begin{equation}
z_{\text{proj}}^{j,n} = \Bigl( \mathbf{R}^j\,\mathbf{X}^{j,n}_{j} + \mathbf{T}^j \Bigr)_z,
\end{equation}
where \((\cdot)_z\) denotes the third (depth) component. Denote the corresponding mono-depth pseudo ground-truth by \(z_{\text{mono}}^{j,n}\). After solving for an optimal scaling factor \(\alpha^*\) to align the two depth maps, the scale-invariant mono-depth loss is defined as
\begin{equation}
L_{\text{depth}} 
= \frac{1}{N}\sum_{n=1}^{N} \Bigl( \alpha^*\, z_{\text{proj}}^{j,n} - z_{\text{mono}}^{j,n} \Bigr)^2, 
\quad \alpha^* = \frac{\sum_{i} z_{\text{proj}}^{j,n}\, z_{\text{mono}}^{j,n}}{\sum_{i}\Bigl(z_{\text{proj}}^{j,n}\Bigr)^2}.
\end{equation}

\vspace{0.5em}
\qheading{3D Self-Consistency.}
Beyond the 2D reprojection losses, we introduce a 3D self-consistency term that aligns the tracking pointmap \(\mathbf{X}^{1}_{j}\) with the reconstruction pointmap \(\mathbf{X}^{j}_{j}\). Let \(\mathcal{I}_1^\prime\) be the set of points in frame \(1\) that remain visible in frame \(j\), and for each point \(n \in \mathcal{I}_1^\prime\), denote its corresponding point (provided by CoTracker) in frame \(j\) by \(n^\prime\). We then penalize the distance between their predicted 3D positions:
\begin{equation}
\mathcal{L}_{\text{align}}
= \sum_{n \in \mathcal{I}_1^\prime}
\bigl\|\mathbf{X}^{1,n}_{j} - \mathbf{X}^{j,n^\prime}_{j}\bigr\|^2.
\end{equation}
Minimizing \(\mathcal{L}_{\text{align}}\) ensures that both branches produce consistent geometry in the same timestamp.

The overall self-supervision loss is given by:
\begin{equation}
\mathcal{L}_{\text{reproj}} = \mathcal{L}_{\text{traj}} + \lambda_1\,\mathcal{L}_{\text{depth}} + \lambda_2\,\mathcal{L}_{\text{align}},
\label{eq:tta_loss}
\end{equation}
with \(\lambda_1\) and \(\lambda_2\) being the weighting factors.
Minimizing \(\mathcal{L}_{\text{reproj}}\) aligns the projected 3D structure with the 2D tracking and monocular depth cues and their 3D self-consistency, enabling unsupervised, target-specific refinement of the 3D geometry and point tracking.

\vspace{0.5em}
\qheading{Test-Time Adaptation.}
To address the gap between synthetic pretraining and real-world data, we incorporate reprojection-based losses to enable test-time adaptation in \modelname. Our framework supports two adaptation paradigms: \textbf{(1) Instance-level adaptation.} During testing, we update \modelname on new sequences using only the aforementioned reprojected losses while freezing the \emph{reconstruction branch}. We freeze these weights because both the 2D trajectories and depth are computed under a purely monocular setting, which does not provide view-alignment supervision. This approach preserves the view-alignment capability captured during pretraining. Moreover, since the pretrained network already encodes strong task-relevant representations, this sequence-specific optimization converges rapidly compared to test-time optimization methods that start from scratch. \textbf{(2) Domain-level adaptation.} Unlike tabula-rasa approaches such as~\cite{wang2024shapeofmotion, lei2024mosca}, which require full re-optimization for each new sequence, \modelname is an end-to-end learning framework that enables test-time adaptation to align the model from its pretraining data distribution to the target video domain. After adapting to a sparse set of target-domain samples, \modelname can directly perform simultaneous reconstruction and tracking on new sequences from the adapted domain without additional optimization.

\section{Experiments}
\modelname performs both dense 3D point tracking and dynamic reconstruction in a unified world coordinate system, all within a single inference. In the following section, we first evaluate our method on 3D tracking and dynamic reconstruction separately, and then present the joint results. 
We also introduce a new benchmark, \textit{WorldTrack}, for 3D tracking in world coordinates, which is not directly covered by previous methods.

\begin{table*}[t]
    \centering
    \renewcommand{\arraystretch}{1.13}
    \renewcommand{\tabcolsep}{6pt}
    \caption{\textbf{World Coordinate 3D Point Tracking.} We report the performance of average points under distance ($\text{APD}_\text{3D}$) after global median alignment. We evaluate the accuracy of both all points and dynamic points. The best results are \textbf{bold}.}
    \vspace{-0.5em}
    \label{tab:3dtracking}
    \resizebox{1\textwidth}{!}{
    \begin{tabular}{@{}llcccccccc@{}}
    \toprule
     & & \multicolumn{4}{c}{All Points} & \multicolumn{4}{c}{Dynamic Points} \\
    \cmidrule(lr){3-6} \cmidrule(lr){7-10}
    \textbf{Category} & \textbf{Methods} & PO & DR & ADT & PStudio & PO & DR & ADT & PStudio \\
    \midrule
    \multirow{2}{*}{\textbf{Combinational}} 
      & SpaTracker+RANSAC-Procrustes 
        & 44.03 & 55.01 & 50.87 & 52.05 & 53.77 & 58.58 & 66.49 & 52.05 \\
      & SpaTracker+MonST3R
        & 47.65 & 55.49 & 51.95 & 50.16 & 58.61 & 59.21 & 69.94 & 50.16 \\
    \midrule
    \multirow{3}{*}{\textbf{Feed-forward}}
      & MonST3R 
        & 33.47 & 58.06 & 74.35 & 51.32 & 39.36 & 51.86 & 67.92 & 51.32 \\
      & SpaTracker 
        & 38.54 & 54.85 & 45.65 & 62.59 & 51.20 & 58.65 & 67.65 & 62.59 \\
        \cmidrule{2-10}
      & \textbf{St4RTrack (Ours)} 
        & \textbf{67.95} & \textbf{73.74} & \textbf{76.01} & \textbf{69.67}
        & \textbf{68.72} & \textbf{68.13} & \textbf{75.34} & \textbf{69.67} \\

    \bottomrule
    \end{tabular}}
\end{table*}

\begin{table}[t]
    \centering
    \footnotesize
    \renewcommand{\arraystretch}{1.25}
    \renewcommand{\tabcolsep}{3pt}
    \caption{\textbf{World Coordinate 3D Reconstruction.}  
    We report performance on both Point Odyssey (PO) and TUM-Dynamics after global median scaling. 
    The best results are in \textbf{bold}.}
    \vspace{-0.5em}
    \label{tab:3drecon}
    \resizebox{\linewidth}{!}{
    \begin{tabular}{@{}llcccc@{}}
    \toprule
        & & \multicolumn{2}{c}{Point Odyssey} & \multicolumn{2}{c}{TUM-Dynamics} \\
    \cmidrule(lr){3-4}\cmidrule(lr){5-6}
    \textbf{Category} & \textbf{Methods} 
        & \textbf{EPE$\downarrow$} & \textbf{APD$\uparrow$} 
        & \textbf{EPE$\downarrow$} & \textbf{APD$\uparrow$} \\
    \midrule
    \multirow{3}{*}{\textbf{w/ Global Align.}}
      & \dus{}+GA        
        & 0.6088 & 43.90 
        & 0.3147 & 70.49 \\
      & MASt3R+GA       
        & 0.4030 & 60.44 
        & 0.5186 & 68.38 \\
      & \mon{}+GA      
        & 0.2629 & 72.31
        & 0.3429 & 63.87 \\
    \midrule
    \multirow{4}{*}{\textbf{Feed-forward}}
      & \dus{}           
        & 0.6386 & 45.79 
        & 0.2891 & 72.26 \\
      & MASt3R           
        & 0.4644 & 56.90 
        & 0.5510 & 66.22 \\
      & \mon{}          
        & 0.3044 & 68.25 
        & 0.3646 & 61.38 \\
        \cmidrule{2-6}
      & \textbf{St4RTrack (Ours)}   
        & \textbf{0.2406} & \textbf{78.73} 
        & \textbf{0.1854} & \textbf{83.42} \\
    \bottomrule
    \end{tabular}}
\end{table}

\subsection{Experimental Details}
\label{sec:exp_details}

\noindent\textbf{Datasets.}
For fully supervised training, we use three synthetic datasets: Point Odyssey (PO)~\cite{zheng2023pointodyssey}, Dynamic Replica (DR)~\cite{karaev2023dynamicstereo}, and Kubric~\cite{greff2022kubric}. All three datasets contain scene and camera motion and provide mesh vertex positions as ground-truth 3D point trajectories. We randomly sample 24 frames with a stride of 1$\sim$6 for each sample sequence. We also filter out less semantically meaningful sequences in PO, resulting in a total of 9.8k sequences for PO, 8.5k for DR, and 5.7k for Kubric dataset.

\vspace{0.5em}
\noindent\textbf{Training and Inference.}
\begin{figure*}   
\includegraphics[width=\linewidth]{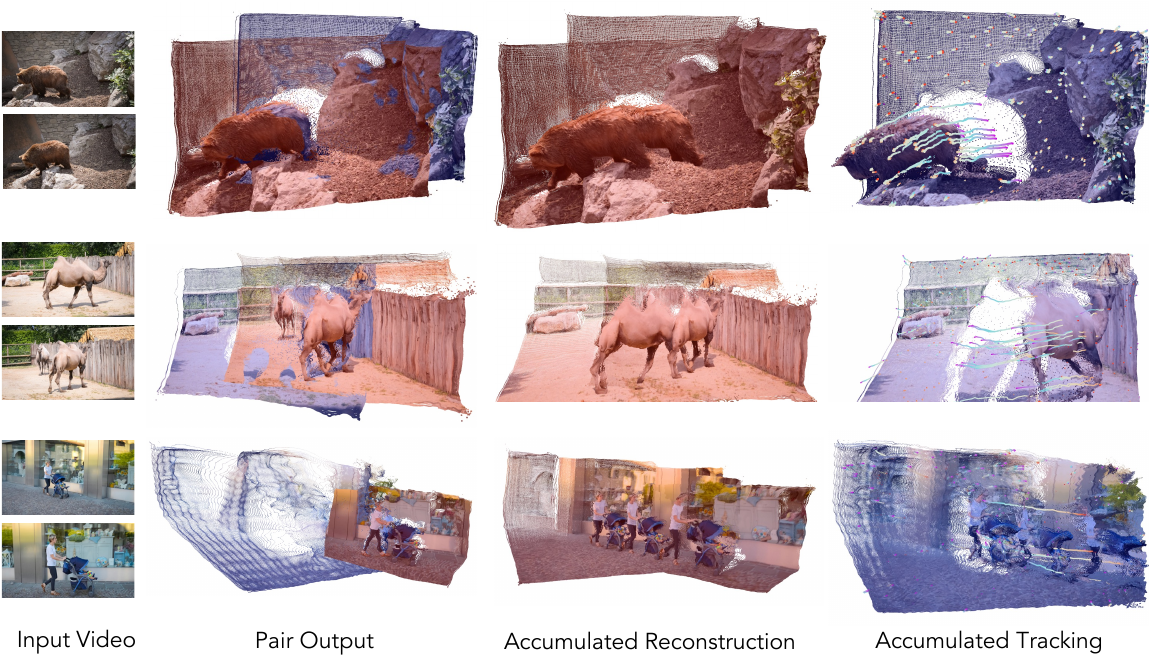}
    \captionof{figure}{\textbf{Qualitative Results.} From left to right, we show our results in \textit{feed-forward inference}: 1) the input video, 2) two pointmaps at frame \(j\) overlayed together, 3) the accumulated reconstruction branch result, and 4) the accumulated tracking branch result. 
    The accumulated reconstruction demonstrates a stable reconstruction of the dynamic scene geometry, while the accumulated tracking illustrates long-term, dense tracking of scene motion.}
    \label{fig: qualitative}
\end{figure*}
\begin{figure*}    \includegraphics[width=\linewidth]{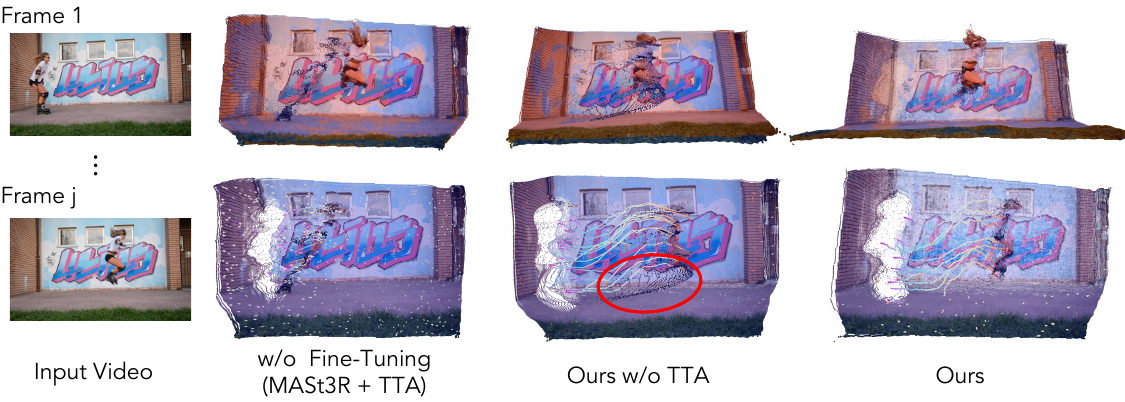}
    \vspace{-1.5em}
    \captionof{figure}
   {\textbf{Ablation Study.} We show the qualitative comparison of our full method and variants that do not pretrain or do not adapt in test time. Predicted pointmaps from two heads are visualized together.}
   \label{fig:ablation}
\end{figure*}
During training, we sample 600 sequences from each dataset per epoch. We use the AdamW optimizer with a learning rate of \(5 \times 10^{-5}\) and a mini-batch size of 1 per GPU. The model is trained for 50 epochs on 4 A100 GPUs, which takes about one day. For test-time adaptation, we run 500 optimization steps on a single sequence, taking approximately 5 minutes on 4 A100 GPUs. At inference time, the model runs at 30 FPS on an RTX 4090. Although the model is trained on sequences of 24 frames, our pair-wise approach allows it to operate on arbitrarily long videos during inference. Refer to~\cref{sec:tta} for more details regarding test-time adaptation. 

\subsection{3D Tracking in World Coordinates}
\paragraph{Datasets.}
3D tracking in world coordinates is a critical aspect that has been largely overlooked by previous benchmarks~\cite{koppula2024tapvid3d}, which are limited to camera coordinate systems. To address this limitation, we propose a new benchmark for 3D tracking in world coordinates. Our benchmark leverages two real-world datasets—Aerial Digital Twin (ADT)~\cite{pan2023ariadigitaltwinnew} and Panoptic Studio~\cite{joo2016panopticstudio}—by converting the TAPVid-3D~\cite{koppula2024tapvid3d} sequences to world coordinates using paired extrinsic parameters. However, it is noteworthy that the limitations of these datasets: the ADT sequences exhibit minimal scene motion, while the Panoptic Studio lacks camera motion. 
To overcome these shortcomings, we include two additional synthetic test sets from Point Odyssey and Dynamic Replica, which have both scene and camera motion. In total, our benchmark comprises four datasets, each containing 50 sequences of 64 frames.

\vspace{0.5em}
\noindent\textbf{Evaluation Metrics.}
We follow the TAPVid-3D protocol and use the \textit{Average percent of Points within Delta (APD)} metric for evaluation. Specifically, we first align the predicted 3D point trajectories with the ground truth by normalizing them with their global median. We then compute the prediction error and measure the percentage of points whose error falls below a given threshold $\delta_{3D}$ (with $\delta_{3D} \in \{0.1\text{m}, 0.3\text{m}, 0.5\text{m}, 1.0\text{m}\}$) over the first 64 frames. Let $\hat{\mathbf{P}}^i_t$ denote the $i$-th predicted point at time $t$ and $\mathbf{P}^i_t$ denote its corresponding ground-truth location. The resulting $\text{APD}_\text{3D}$ is then computed as follows:
\begin{equation}
\text{APD}_\text{3D} \equiv \sum_{i,t} \mathds{1}\Bigl(\|\hat{\mathbf{P}}^i_t - \mathbf{P}^i_t\| < \delta_{3D}\Bigr),
\label{eq:apd}
\end{equation}
where \(\mathds{1}(\cdot)\) is the indicator function and \(\|\cdot\|\) denotes the Euclidean norm.

\vspace{0.5em}
\noindent\textbf{Baselines.}
Since no existing work explicitly performs 3D tracking in world coordinates, for our feedforward baselines, we compare against the camera coordinate 3D tracking method SpatialTracker~\cite{xiao2024spatialtracker} and a dynamic 3D reconstruction method \mon{}~\cite{zhang2025monst3r} (as a non-tracking baseline). 
In addition, we implement two combinational baselines for world coordinate 3D tracking. The first baseline applies Procrustes alignment~\cite{umeyama1991least} and RANSAC~\cite{fischler1981random} to the camera coordinate 3D tracks predicted by SpatialTracker to offset the camera motion. The second baseline leverages the camera poses predicted by the dynamic SLAM method \mon{} to compensate for camera motion.

\vspace{0.5em}
\noindent\textbf{Results.}
As shown in~\cref{tab:3dtracking}, we achieve state-of-the-art performance, with test-time adaptation proving particularly beneficial for dynamic points. Notably, on the Panoptic Studio dataset, which is captured with a fixed camera and can be considered a fair benchmark for camera coordinate tracking methods, our approach still outperforms SpatialTracker~\cite{xiao2024spatialtracker}.
It is worth noting that although our model is trained on sequences of 24 frames, it generalizes well to longer sequences, including 64-frame videos. Refer to~\cref{sec:more_results} for more results.

\subsection{Dynamic 3D Reconstruction}

\noindent\textbf{Datasets.}
We evaluate on both synthetic and real-world data. For the synthetic data, we use Point Odyssey~\cite{zheng2023pointodyssey}. For real-world evaluation, we employ TUM-Dynamics~\cite{tum}, a subset of a SLAM dataset featuring moving people, dense depth maps, and accurate camera poses. 

\vspace{0.5em}
\noindent\textbf{Evaluation Metrics.}
Unlike prior works~\cite{wang2025cut3r,zhang2025monst3r} that separately evaluate video depth and camera pose estimation, we directly compare the reconstructed 3D point clouds to the ground truth using the Average percent of Points within Distance (APD) and End-Point Error (EPE) metrics. We filter out ambiguous floating points in the ground truth data and align the point clouds for each sequence using the median scale before evaluation. 

\vspace{0.5em}
\noindent\textbf{Baselines.}
We compare our method against \mon{}, MASt3R and \dus{}, both with global alignment (w/ GA) and in feedforward mode. For the feedforward baselines, we construct image pairs of a video in the form of \modelname, that align all frames to a common anchor frame.

\vspace{0.5em}
\noindent\textbf{Results.}
As in~\cref{tab:3drecon}, we also achieve state-of-the-art performance on the task of 3D reconstruction in the world coordinate. Although \mon{} is designed for 3D reconstruction for dynamic scenes, it still underperforms \modelname even with global alignment. This further highlights the benefit of jointly tracking and reconstruction.  Since we freeze the reconstruction head to preserve the 3D prior, 3D reconstruction results are similar with test-time adaptation. 

\subsection{Joint Tracking and Reconstruction in the World}

Our method simultaneously predicts 3D point trajectories and 3D point clouds in a single feed-forward pass, which we evaluate separately in previous sections.
In this section, we present qualitative results that visualize both the raw 3D point trajectories and 3D point clouds within the same world coordinates, as shown in Fig.~\ref{fig: qualitative}. 
The ``pair output" result demonstrates that the outputs from the tracking and reconstruction branches align well at the same time step. Additionally, the accumulated reconstruction indicates consistency in static regions, while the accumulated tracking shows that our method estimates accurate and smooth 3D tracks over time.

\subsection{Ablation Study}

We perform an ablation study to evaluate two key design choices of our method and present qualitative results in Fig.~\ref{fig:ablation}. First, we assess the effectiveness of our pretraining stage by directly applying test-time adaptation to a pretrained checkpoint from \mon{}~\cite{zhang2025monst3r}, without finetuning the base model on our training datasets. As shown in Fig.~\ref{fig:ablation} (column 2), the baseline exhibits unaligned pointmaps between the tracking and reconstruction branches, underscoring the importance of pretraining on synthetic data—even in the presence of a domain gap with real-world data.

Second, we evaluate the impact of our proposed test-time adaptation. As demonstrated in Fig.~\ref{fig:ablation} (column 3), the adapted model successfully corrects drifting points, ensuring that points consistently trace back to their original spatial locations in the first frame. This finding supports our analysis that small-scale training data alone is insufficient for fine-grained prediction, particularly at the boundaries of moving objects. In contrast, \modelname produces spatially aligned pointmaps with significantly fewer drifting points. The colorful tails in the visualization indicate the long-term trajectories, while the accurately predicted geometry in dynamic regions results in a crisp and precise rendering.

\section{Discussion}
Despite \modelname presents a promising step toward a unified understanding of dynamic scene geometry and motion in a minimalist way, a challenge arises from the per-frame setting. In particular, issues such as scale misalignment, large camera movements, and occlusions are not fully resolved. Incorporating temporal attention across multiple frames would help capture richer motion priors and alleviate these limitations. Another limitation arises from the pretraining dataset’s limited diversity and realism in both geometry and motion, necessitating test‐time adaptation to improve \modelname’s robustness in out‐of‐distribution scenarios. However, it still struggles with highly complex motions. Expanding the training set is therefore a key direction for future work. We envision that large-scale pretraining, when compute permits, could significantly boost \modelname's performance and enable it to better handle complex, in-the-wild videos.

\section{Conclusion}
We introduce \modelname, a feed-forward framework that \textit{simultaneously} achieves 3D point tracking and dynamic reconstruction \emph{in the world coordinate} from monocular videos using a unified representation. Alongside, we present a novel benchmark, \textit{WorldTrack}, for systematically evaluating dynamic 3D scene geometry and motion estimation in a global reference frame. Our method achieves state-of-the-art performance on both synthetic and real-world datasets, while also extending beyond fully supervised paradigms by enabling test-time adaptation.

\section{Acknowledgements}
 We would like to thank Aleksander Holynski, Yifei Zhang, Chung Min Kim, and Brent Yi for helpful discussions. We especially thank Aleksander Holynski for his guidance and feedback.

{
    \small
    \bibliographystyle{ieeenat_fullname}
    \bibliography{main}
}

\clearpage
\maketitlesupplementary

\tableofcontents

\appendix

\section{Differentiable Camera Pose Estimation}
\label{sec:diff_cam}

We seek to backpropagate the projection loss to the 3D pointmaps through the camera pose. To this end, we build upon the RANSAC-PnP approach from \dus{}~\cite{Wang2023DUSt3RG3}, which initially solves for pose \(\mathbf{P}^*\) (rotation and translation) by matching per-pixel 2D-3D correspondences in the reconstruction pointmap \(\mathbf{X}^j_j\). However, RANSAC is inherently non-differentiable.

To enable end-to-end gradients, we adopt the derivative-based Gauss-Newton (GN) solver inspired by EPro-PnP~\cite{chen2023epro}. Specifically, after obtaining a \emph{detached} solution \(\mathbf{P}^*\) from RANSAC-PnP, we refine it using one GN step:
\begin{equation}
    \Delta \mathbf{P} \;=\; -\bigl(J^\top J\bigr)^{-1}\,J^\top\,F(\mathbf{P}^*),
    \label{gnstep}
\end{equation}
where \(F(\mathbf{P}^*) = [\,f_1^\top(\mathbf{P}^*),\,\dots,\,f_N^\top(\mathbf{P}^*)]^\top\) is the flattened reprojection error for all \(N\) points, and \(J = \tfrac{\partial F(\mathbf{P})}{\partial \mathbf{P}}\bigl\rvert_{\mathbf{P}=\mathbf{P}^*}\) is its Jacobian. The term \(J^\top J\) approximates the Hessian of the negative log-likelihood (NLL), while \(J^\top F(\mathbf{P}^*)\) is the gradient of the NLL with respect to the pose. This gradient effectively \emph{pushes} the incremental solution $\Delta \mathbf{P}$ toward reducing the reprojection errors. The final \emph{differentiable} pose estimate is:
\begin{equation}
\mathbf{P} \;=\; \mathbf{P}^* \;+\; \Delta \mathbf{P}.
\label{regloss}
\end{equation}
Since \(\mathbf{P}^*\) is detached, only the GN increment \(\Delta \mathbf{P}\) remains differentiable, allowing the reprojection loss to backpropagate through \(\mathbf{P}\) and thus refine the 3D pointmaps.

\begin{table*}[t]
    \centering
    \footnotesize
    \renewcommand{\arraystretch}{1.25}
    \renewcommand{\tabcolsep}{6pt}
    \caption{\textbf{World Coordinate 3D Point Tracking (EPE - Global Median) .}  
    We report end‑point error (EPE; lower is better) for both all points and dynamic points
    after global median alignment.  The best (lowest) values are in \textbf{bold}.}
    \vspace{-0.5em}
    \label{tab:3dtracking_epe}
    \resizebox{\textwidth}{!}{
    \begin{tabular}{@{}llcccccccc@{}}
    \toprule
        & & \multicolumn{4}{c}{All Points} & \multicolumn{4}{c}{Dynamic Points} \\
    \cmidrule(lr){3-6}\cmidrule(lr){7-10}
    \textbf{Category} & \textbf{Methods}
        & PO & DS & ADT & PStudio
        & PO & DS & ADT & PStudio \\
    \midrule
    \multirow{2}{*}{\textbf{Combinational}}
        & SpaTracker+RANSAC-Procrustes
            & 0.6408 & 0.9185 & 0.5876 & 0.4266
            & 0.4358 & 1.0444 & 0.1600 & 0.4266 \\
        & SpaTracker+MonST3R
            & 0.5917 & 0.8823 & 0.5362 & 0.4837
            & 0.4085 & 0.9136 & 0.1511 & 0.4837 \\
    \midrule
    \multirow{3}{*}{\textbf{Feed‑forward}}
        & MonST3R
            & 0.9021 & 0.4387 & 0.2721 & 0.4568
            & 0.6452 & 0.5313 & 0.1578 & 0.4568 \\
        & SpaTracker
            & 0.7499 & 0.9274 & 0.8530 & 0.3094
            & 0.4695 & 1.0828 & 0.1628 & 0.3094 \\
        \cmidrule{2-10}
        & \textbf{Ours}
            & \textbf{0.3140} & \textbf{0.2682} & \textbf{0.2680} & \textbf{0.2637}
            & \textbf{0.2970} & \textbf{0.2961} & \textbf{0.1212} & \textbf{0.2637} \\
    \bottomrule
    \end{tabular}}
\end{table*}

\begin{table*}[t]
    \centering
    \footnotesize
    \renewcommand{\arraystretch}{1.5}
    \renewcommand{\tabcolsep}{2pt}
    \caption{\textbf{World Coordinate 3D Point Tracking (APD/EPE - SIM(3)).}  
    Each cell shows APT\textsubscript{3D} (higher is better) \emph{/} EPE (lower is better) after global IM(3) alignment.  
    The best APT (highest) and the best EPE (lowest) in every column are \textbf{bold}.}
    \vspace{-0.5em}
    \label{tab:3dtracking_sim3}
    \resizebox{\textwidth}{!}{
    \begin{tabular}{@{}llcccccccc@{}}
    \toprule
        & & \multicolumn{4}{c}{All Points} & \multicolumn{4}{c}{Dynamic Points} \\
    \cmidrule(lr){3-6}\cmidrule(lr){7-10}
    \textbf{Category} & \textbf{Methods}
        & PO & DR & ADT & PStudio
        & PO & DR & ADT & PStudio \\
    \midrule
    \multirow{2}{*}{\textbf{Combinational}}
        & SpaTracker+Procrustes\
            & 46.20/0.5670 & 55.10/0.5292 & 59.40/0.4027 & 67.82/0.2660
            & 61.00/0.3338 & 61.65/0.3720 & \textbf{88.65}/0.0596 & 67.82/0.2660 \\
        & SpaTracker+MonST3R
            & 48.23/0.5388 & 56.78/0.5069 & 60.01/0.3910 & 64.32/0.2971
            & 61.78/0.3290 & 61.88/0.3681 & 87.32/\textbf{0.0485} & 64.32/0.2971 \\
    \midrule
    \multirow{3}{*}{\textbf{Feed‑forward}}
        & MonST3R
            & 37.62/0.8073 & 64.83/0.3725 & 79.48/0.1881 & 64.11/0.3015
            & 48.95/0.4768 & 55.36/0.3872 & 84.73/0.0720 & 64.11/0.3015 \\
        & SpaTracker
            & 43.17/0.6079 & 54.65/0.5324 & 53.96/0.4963 & \textbf{80.76}/\textbf{0.1650}
            & 60.49/0.3374 & 61.32/0.3750 & 87.68/0.0616 & \textbf{80.76}/\textbf{0.1650} \\
        \cmidrule{2-10}
        & \textbf{Ours}
            & \textbf{71.84}/\textbf{0.2774} & \textbf{76.28}/\textbf{0.2436} & \textbf{83.03}/\textbf{0.1631} & 76.97/0.1969
            & \textbf{67.43}/\textbf{0.2870} & \textbf{67.90}/\textbf{0.2627} & 85.34/0.0688 & 76.97/0.1969 \\
    \bottomrule
    \end{tabular}}
\end{table*}

\begin{table}[t]
    \centering
    \footnotesize
    \renewcommand{\arraystretch}{1.25}
    \renewcommand{\tabcolsep}{3pt}
    \caption{\textbf{World Coordinate 3D Reconstruction (APD/EPE - SIM(3)).}  
    Results on Point Odyssey (PO) and TUM‑Dynamics after global SIM(3) alignment.  
    Lower is better for EPE, higher is better for APT.  The best results are in \textbf{bold}.}
    \vspace{-0.5em}
    \label{tab:3drecon_sim3}
    \resizebox{\linewidth}{!}{
    \begin{tabular}{@{}llcccc@{}}
    \toprule
        & & \multicolumn{2}{c}{Point Odyssey} & \multicolumn{2}{c}{TUM‑Dynamics} \\
    \cmidrule(lr){3-4}\cmidrule(lr){5-6}
    \textbf{Category} & \textbf{Methods}
        & \textbf{EPE$\downarrow$} & \textbf{APT$\uparrow$}
        & \textbf{EPE$\downarrow$} & \textbf{APT$\uparrow$} \\
    \midrule
    \multirow{3}{*}{\textbf{w/ Global Align.}}
      & \dus{}+GA
        & 0.3541 & 62.42
        & 0.2989 & \textbf{69.23} \\
      & MASt3R+GA
        & 0.3717 & 61.31
        & 0.5294 & 49.81 \\
      & \mon{}+GA
        & \textbf{0.2601} & 69.31
        & 0.3173 & 66.00 \\
    \midrule
    \multirow{4}{*}{\textbf{Feed‑forward}}
      & \dus{}
        & 0.4251 & 56.70
        & 0.3092 & 67.48 \\
      & MASt3R
        & 0.4473 & 55.09
        & 0.5862 & 45.43 \\
      & \mon{}
        & 0.3462 & 62.10
        & 0.3508 & 62.83 \\
      & \textbf{\modelname}
        & 0.2741 & \textbf{69.53}
        & \textbf{0.2413} & \textbf{74.14} \\
    \bottomrule
    \end{tabular}}
\end{table}

\section{Details on the \textit{WorldTrack} Benchmark}
\subsection{Datasets}
\qheading{Dataset Preparation.} For the two real-world datasets, we adopt the 3D camera coordinate tracking annotation of ADT and Panoptic Studio from the TAPVID-3D dataset. Using the paired camera parameters provided, we transform the camera coordinates to the world coordinate system. For the two synthetic datasets, we use the test sets from Point Odyssey and Dynamic Replica Dataset. We uniformly downsample the query points to approximately 1,000 per sequence. Each sequence contains 128 sampled frames, though only the first 64 frames are used for evaluation. This results in 160 and 140 sequences from Point Odyssey and Dynamic Replica, respectively. From these, we randomly sample 50 sequences per dataset for evaluation.

\vspace{0.5em}
\qheading{Filtering Criteria.} To ensure data quality, we apply several filtering strategies: For TUM, we keep the pixels which associate with depth values within 0.1 - 5 meters, as the depth camera is less accurate at long range. For Point Odyssey, we exclude sequences generated in the Kubric style~\cite{greff2022kubric} due to their lack of realism. We also remove scenes with ambiguous depth (e.g., heavy foggy conditions), and any frames where the camera intrinsics are dynamic.

\subsection{Additional Quantitative Evaluation}
\label{sec:more_results}
Following TAPVid-3D~\cite{koppula2024tapvid3d}, we adopt global median scale alignment, since both our predictions and the ground truth use the first frame’s camera coordinate system as the world coordinate. The Average Percent of Points within Distance (APD$_\text{3D}$) measures the overall accuracy of the 3D trajectories in world coordinates, while Euclidean endpoint error (EPE) offers a complementary perspective on localization accuracy. Accordingly, we additionally report EPE results on the WorldTrack benchmark. As shown in Table~\ref{tab:3dtracking_epe}, \modelname attains state-of-the-art EPE on all sub-test sets, consistent with the APD$_\text{3D}$ results in the main paper.

Beyond alignment to the first camera’s pose, we also evaluate under SIM(3) alignment (i.e., SE(3) plus a global scale factor) for both APD$_\text{3D}$ and EPE to assess performance of 3D tracking (See~\cref{tab:3dtracking_sim3}) and reconstruction (See~\cref{tab:3drecon_sim3}) under a more flexible registration. Comprehensive evaluations show that \modelname achieves state-of-the-art performance in most scenarios.

\subsection{Qualitative Evaluation}
\begin{figure*}   
\includegraphics[width=\linewidth]{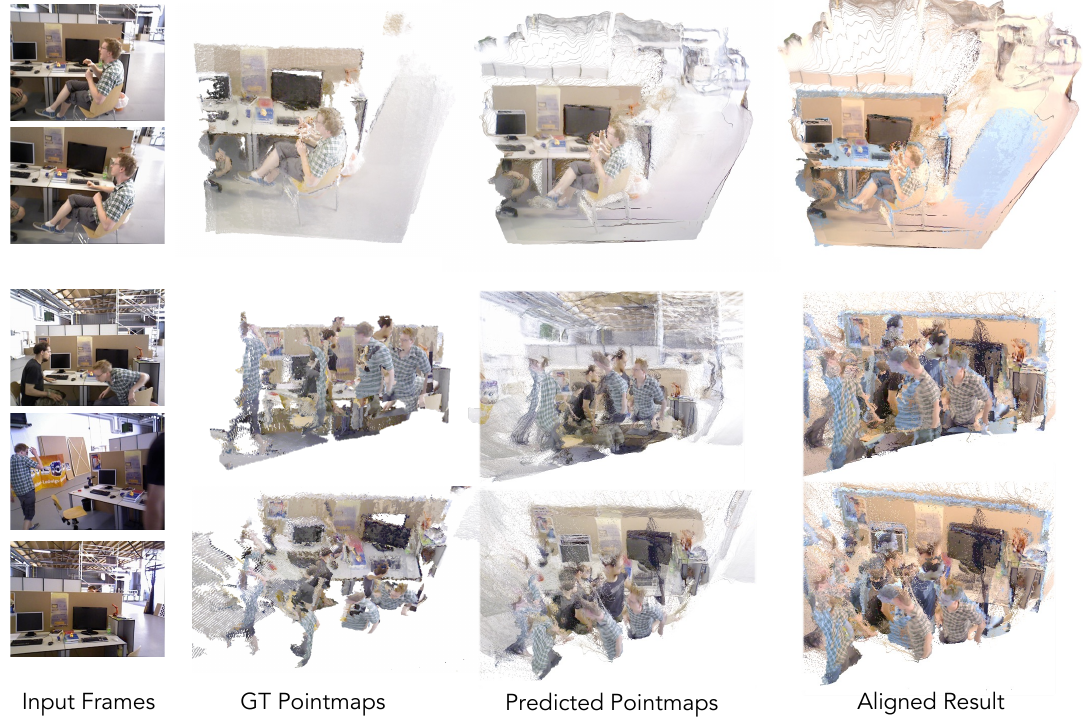}
    \captionof{figure}{\textbf{Reconstruction Results of St4RTrack on TUM-Dynamics Dataset.} From left to right, we show 1) the sampled frames from the input sequence of 64 frames, 2) the subsampled ground truth pointmaps, 3) the predicted pointmaps of our method, and 4) the aligned results of the predicted and GT pointmaps with median-scale. Note that the reconstruction result is inferred in a feed-forward way.}
    \label{fig: bench_recon1}
\end{figure*}

\begin{figure*}   
\includegraphics[width=\linewidth]{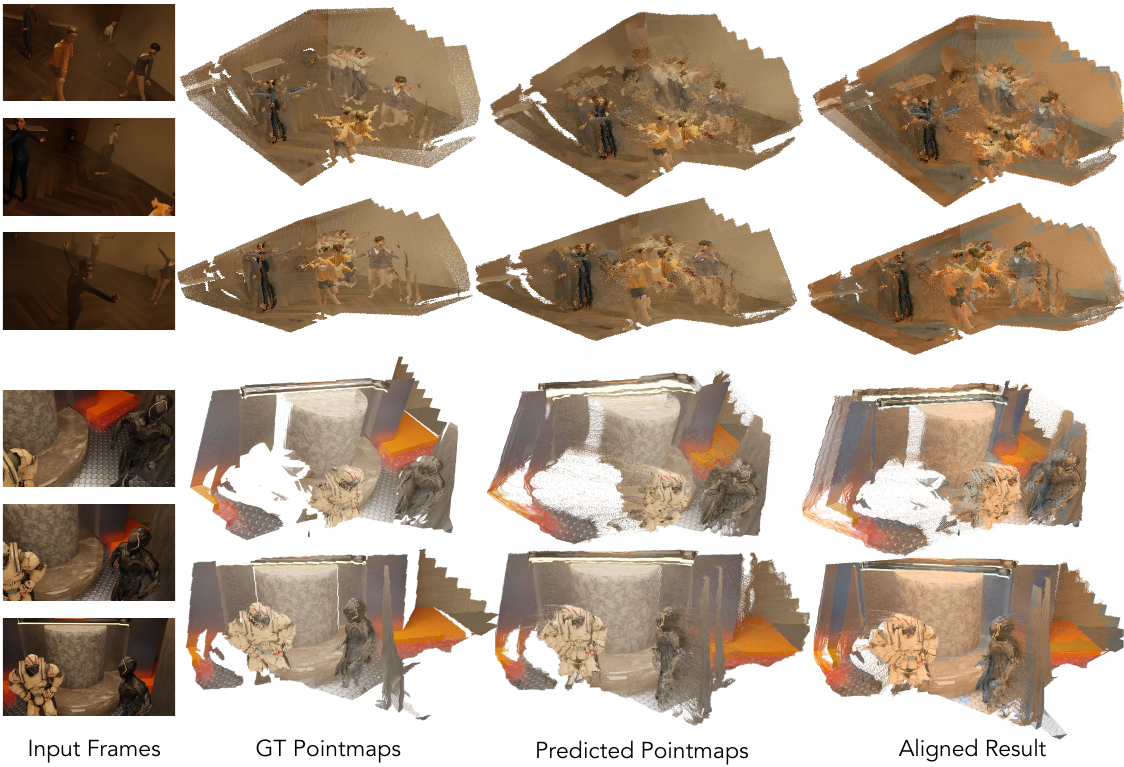}
    \captionof{figure}{\textbf{Reconstruction Results of St4RTrack on Point Odyssey Dataset.} From left to right, we show 1) the sampled frames from the input sequence of 64 frames, 2) the subsampled ground truth pointmaps, 3) the predicted pointmaps of our method, and 4) the aligned results of the predicted and GT (yellow) pointmaps with median-scale. Note that the reconstruction result is inferred in a feed-forward way.}
    \label{fig: bench_recon2}
\end{figure*}

\begin{figure*}   
\includegraphics[width=\linewidth]{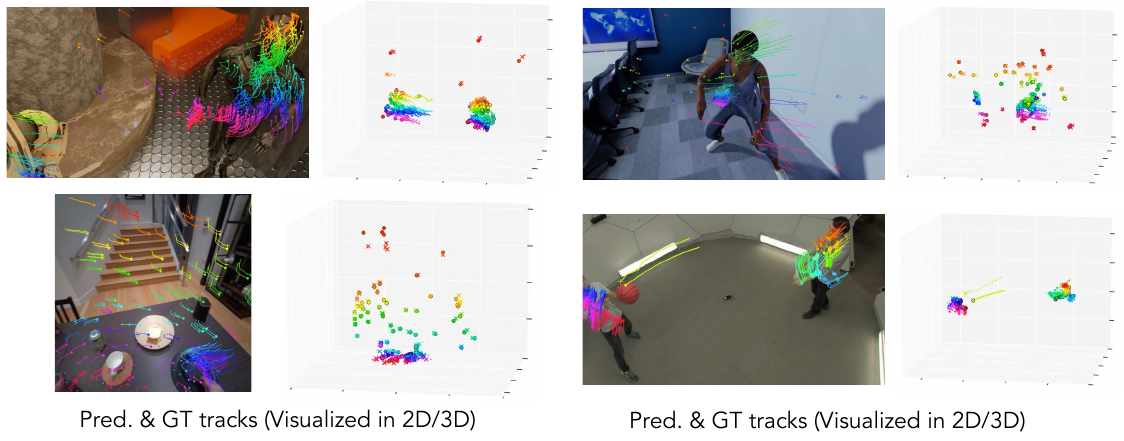}
    \captionof{figure}{\textbf{Tracking Results of St4RTrack on WorldTrack Benchmark.} We show the results of the predicted tracks aligned with the ground truth tracks, visualized in 2D and 3D. The corresponding datasets are Point Odyssey (top left), Dynamic Replica (top right), Arial Digital Twin (bottom left), and Pnapotic Studio (bottom right).}
    \label{fig: bench_track}
\end{figure*}

We present the qualitative results of our fully feed-forward approach on WorldTrack benchmark. Specifically, we show the reconstruction results in~\cref{fig: bench_recon1} (TUM-Dynamics) and \cref{fig: bench_recon2} (Point Odyssey). We show the tracking results of four datasets in~\cref{fig: bench_track}.

\section{Details of Test-Time Adaptation}
\label{sec:tta}

\begin{table}[t]
    \centering
    \footnotesize
    \renewcommand{\arraystretch}{1.25}
    \renewcommand{\tabcolsep}{4pt}
    \caption{\textbf{World Coordinate 3D Tracking (Median‑Scale).}  
    End‑point error (EPE ↓) and APT\textsubscript{3D} ↑ for DR and PStudio after global median scaling.  
    Best (lowest EPE / highest APT\textsubscript{3D}) in each column is shown in \textbf{bold}.}
    \vspace{-0.5em}
    \label{tab:wct_tracking_ablation}
    \resizebox{\linewidth}{!}{
    \begin{tabular}{@{}lcccc@{}}
    \toprule
        & \multicolumn{2}{c}{DR} & \multicolumn{2}{c}{PStudio} \\
    \cmidrule(lr){2-3}\cmidrule(lr){4-5}
    \textbf{Methods}
        & \textbf{EPE$\downarrow$} & \textbf{APT$\uparrow$}
        & \textbf{EPE$\downarrow$} & \textbf{APT$\uparrow$} \\
    \midrule
    Spatialtracker+Procrustes‑RANSAC           & 0.9185 & 55.01 & 0.4266 & 52.05 \\
    St4RTrack                                  & 0.2682 & 73.74 & 0.2637 & 69.67 \\
    St4RTrack + TTA (per‑sequence)             & \textbf{0.2472} & \textbf{76.07} & \textbf{0.2243} & \textbf{73.71} \\
    \hline
    St4RTrack + TTA (per‑dataset)              & 0.2547 & 74.86 & 0.2280 & 73.30 \\
    \quad w/o trajectory loss                  & 0.2767 & 72.75 & 0.2421 & 72.50 \\
    \quad w/o depth loss                       & 0.5524 & 48.22 & 0.2975 & 66.50 \\
    \quad w/o alignment loss                   & 0.3263 & 66.65 & 0.3357 & 60.07 \\
    \quad w/o pre‑training                     & 0.3377 & 65.50 & 0.3801 & 57.71 \\
    \bottomrule
    \end{tabular}}
\end{table}

\subsection{Implementation setup}
 We set the weights of different loss factor in \cref{eq:tta_loss} to $\lambda_{\text{traj}}=1$, $\lambda_{\text{depth}}=10$, and $\lambda_{\text{align}}=5$. For WorldTrack evaluation, the two test-time adaptations are set up as the following, \textbf{Sequence-Level (Instance) Adaptation:} Fine-tune a separate model for each of the 50 sequences. We sample 300 frames per epoch, train for 3 epochs, and use a batch size of 4. \textbf{Dataset-Level (Domain) Adaptation:} Fine-tune a single model on the entire dataset. We sample 100 frames per epoch, train for 15 epochs, and use a batch size of 4.

\subsection{Ablation Studies}
We ablate (1) the performance gain from the feed‑forward \modelname, instance‑level adaptation, and domain‑level adaptation, and (2) the contribution of each TTA component by omitting individual elements. Table~\ref{tab:wct_tracking_ablation} summarizes our findings. First, both TTA variants yield substantial improvements over the feed‑forward mode, with instance‑level adaptation achieving the highest accuracy, as it fully specializes to each test sequence. Second, removing any single TTA component—trajectory loss, depth loss, alignment loss, or synthetic pretraining—causes a performance drop in all scenarios, underscoring the necessity of each component.

\section{Additional Results}
\begin{figure*}  
\centering
\vspace{-3em}
\includegraphics[width=0.9\linewidth]{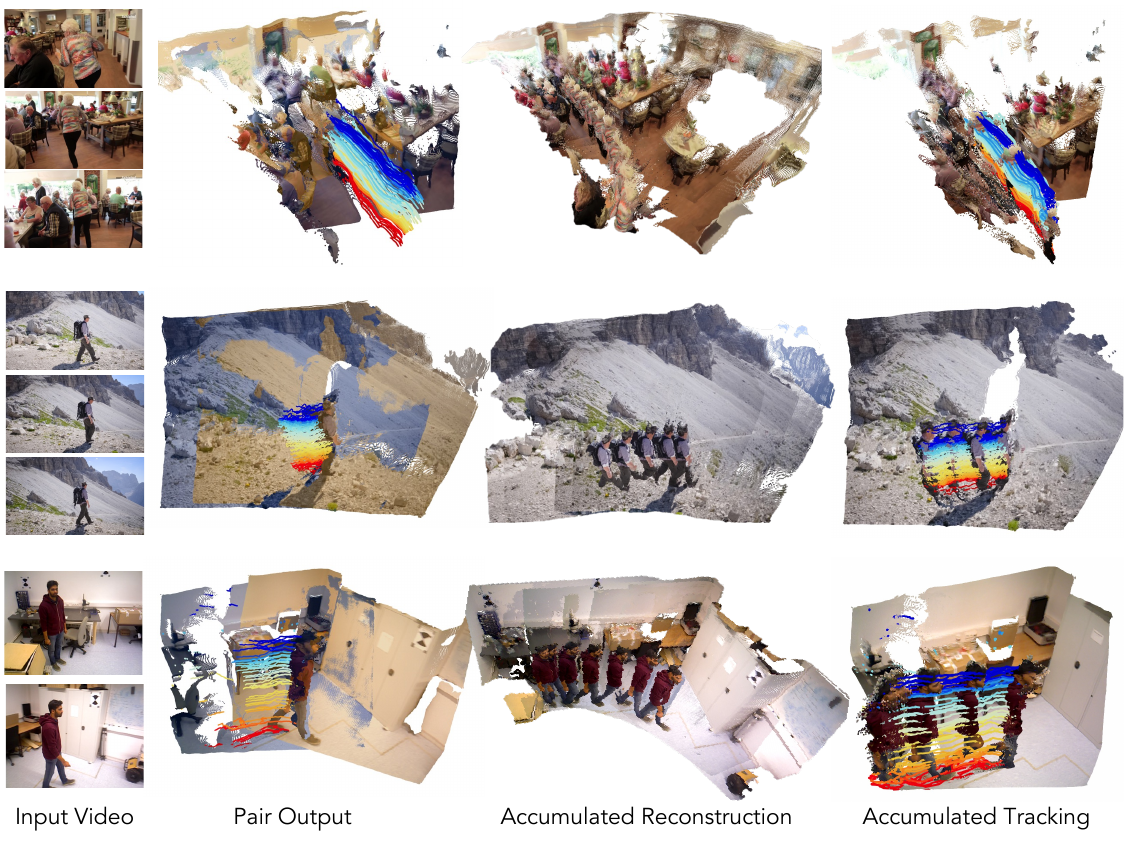}
    \captionof{figure}{\textbf{Fully \textit{Feed-Forward} Inference Results of St4RTrack.} We show from left to right: 1) the input video, 2) the pairwise output for tracking (in blue) and reconstruction (in yellow) of the same frame, 3) the accumulated results of the reconstruction pointmaps, and 4) the accumulated results of the tracking pointmaps. Note that we anchor the \textit{middle frame} as the reference frame for point tracking.}
    \label{fig: ffw_results}
\end{figure*}

\begin{figure*}  
\centering
\includegraphics[width=0.9\linewidth]{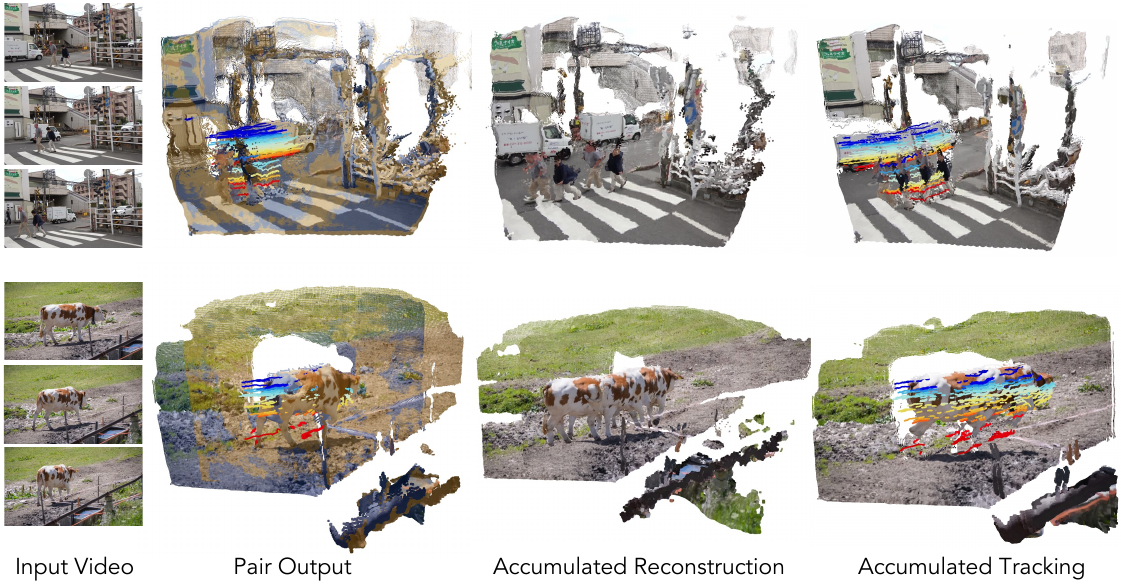}

    \captionof{figure}{\textbf{Test-Time Adaptation Results of St4RTrack.} From left to right, we show 1) the input video, 2) the pairwise output for tracking (in blue) and reconstruction (in yellow) of the same frame, 3) the accumulated results of the reconstruction pointmaps, and 4) the accumulated results of the tracking pointmaps.}
    \label{fig: tta_results}
\end{figure*}

Below, we present additional qualitative results for both feed‑forward inference (\cref{fig: ffw_results}) and test‑time adaptation (\cref{fig: tta_results}).

\label{sec:ffi_result}

\label{sec:tta_result}

\end{document}